\pdfoutput=1

\documentclass[11pt]{article}

\usepackage{acl}

\usepackage{times}
\usepackage{latexsym}
\usepackage{graphicx}
\usepackage{amsmath}
\usepackage{hyperref}

\usepackage[T1]{fontenc}

\usepackage[utf8]{inputenc}

\usepackage{microtype}

\usepackage{listings}
\usepackage{xcolor}
\definecolor{codered}{rgb}{1,0,0}
\definecolor{codegreen}{rgb}{0,0.6,0}
\definecolor{codegray}{rgb}{0.5,0.5,0.5}
\definecolor{codepurple}{rgb}{0.58,0,0.82}
\definecolor{backcolour}{rgb}{0.95,0.95,0.92}

\lstdefinestyle{mystyle}{
  backgroundcolor=\color{backcolour}, commentstyle=\color{codegreen},
  keywordstyle=\color{magenta},
  numberstyle=\tiny\color{codegray},
  stringstyle=\color{codepurple},
  basicstyle=\ttfamily\footnotesize,
  breakatwhitespace=false,         
  breaklines=true,                 
  captionpos=b,                    
  keepspaces=true,                 
  numbers=left,                    
  numbersep=5pt,                  
  showspaces=false,                
  showstringspaces=false,
  showtabs=false,                  
  tabsize=2
}

\title{Text Smoothing: Enhance Various Data Augmentation Methods on Text Classification Tasks}

\author{
    Xing Wu\textsuperscript{\rm 1,2,3}, Chaochen Gao\textsuperscript{\rm 1,2}\thanks{Work done during internship at Kuaishou Inc. The first two authors contribute equally.}, Meng Lin\textsuperscript{\rm 1,2}, Liangjun Zang\textsuperscript{\rm 1}, Zhongyuan Wang\textsuperscript{\rm 3}, Songlin Hu\textsuperscript{\rm 1,2}
    \\
    \textsuperscript{\rm 1}Institute of Information Engineering, Chinese Academy of Sciences, Beijing, China\\
    \textsuperscript{\rm 2}School of Cyber Security, University of Chinese Academy of Sciences, Beijing, China\\
    \textsuperscript{\rm 3}Kuaishou Technology, Beijing, China
    \\
    \{gaochaochen,linmeng,zangliangjun,husonglin\}@iie.ac.cn
    \\
    \{wuxing,wangzhongyuan\}@kuaishou.com
}

\begin{document}
\maketitle
\begin{abstract}
Before entering the neural network, a token is generally converted to the corresponding one-hot representation, which is a discrete distribution of the vocabulary. 
Smoothed representation is the probability of candidate tokens obtained from a pre-trained masked language model, which can be seen as a more informative substitution to the one-hot representation. 
We propose an efficient data augmentation method, termed \textbf{text smoothing}, by converting a sentence from its one-hot representation to a controllable smoothed representation.
We evaluate text smoothing on different benchmarks in a low-resource regime. 
Experimental results show that text smoothing outperforms various mainstream data augmentation methods by a substantial margin. Moreover, text smoothing can be combined with those data augmentation methods to achieve better performance.
\end{abstract}

\section{Introduction}
Data augmentation is a widely used technique, especially in the low-resource regime. It increases the size of the training data to alleviate overfitting and improve the robustness of deep neural networks.
In the field of natural language processing (NLP), various data augmentation techniques have been proposed. 
One most commonly used method is to randomly select tokens in a sentence and replace them with semantically similar tokens to synthesize a new sentence \cite{wei2019eda, kobayashi2018contextual, wu2019conditional}.
\cite{kobayashi2018contextual} proposes contextual augmentation to predict the probability distribution of replacement tokens by using the LSTM language model and sampling the replacement tokens according to the probability distribution.
\cite{wu2019conditional} uses BERT's \cite{devlin2018bert} masked language modeling (MLM) task to extend contextual augmentation by considering deep bi-directional context. \cite{kumar2020data} further propose to use different types of transformer based pre-trained models for conditional data augmentation in the low-resource regime.

\begin{figure}[!tbp]
\centering
\includegraphics[width=0.45\textwidth,height=0.2\textheight]{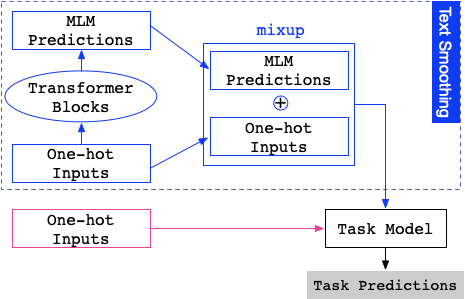}
\caption{The blue part demonstrates the use of text smoothing data augmentation for downstream tasks, and the red part directly uses the original input.}
\label{fig:ts_1}
\vspace{-0.4cm}
\end{figure}

MLM takes masked sentences as input, and typically 15\% of the original tokens in the sentences will be replaced by the [MASK] token.
Before entering MLM, each token in sentences needs to be converted to its one-hot representation, a vector of the vocabulary size with only one position is 1 while the rest positions are 0. 
MLM outputs the probability distribution of the vocabulary size of each mask position. Through large-scale pre-training, it is expected that the probability distribution is as close as possible to the ground-truth one-hot representation.
Compared with the one-hot representation, the probability distribution predicted by pre-trained MLM is a ``smoothed'' representation, which can be seen as a set of candidate tokens with different weights. Usually, most of the weights are distributed on contextual-compatible tokens. 
Multiplying the smooth representation by the word embedding matrix can obtain a weighted summation of the word embeddings of the candidate words, termed smoothed embedding, which is more informative and context-rich than the one-hot's embedding obtained through lookup operation. Therefore, the use of smoothed representation instead of one-hot representation as the input of the model can be 
seen as an efficient weighted data augmentation method.
To get the smoothed representation of all the tokens of the entire sentence with only one forward process in MLM, we do not explicitly mask the input. Instead, we turn on the dropout of MLM and dynamically randomly discard a portion of the weight and hidden state at each layer.

\begin{figure}[!tbp]
\centering
\includegraphics[width=0.5\textwidth,height=0.2\textheight]{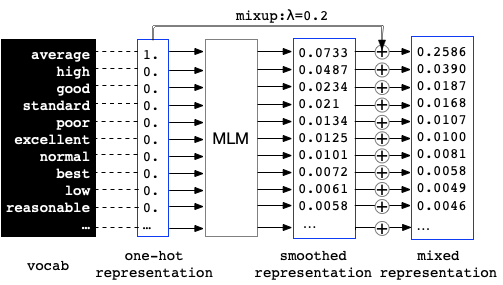}
\caption{Interpolation of the smoothed representation and the original one-hot representation.}
\label{fig:ts_3}
\vspace{-0.4cm}
\end{figure}

An unneglectable situation is that some tokens appear more frequently than others in similar contexts during pre-training, which will cause the model to have a preference for these tokens. This is harmful for downstream tasks such as fine-grained sentiment classification. For example, given ``The quality of this shirt is average .", the ``average" token is most relevant to the label.
The smoothed representation through the MLM at the position of ``average" is shown in Figure \ref{fig:ts_3}. Although the probability of ``average" is the highest, more probabilities are concentrated on tokens conflict with the task label, such as ``high", ``good" or ``poor''. Such a smoothed representation is hardly a good augmented input for the task.
To solve this problem, \cite{wu2019conditional} proposed to train label embedding to constraint MLM predict label compatible tokens.
However, under the condition of low resources, it is not easy to have enough label data to provide supervision.
We get inspiration from the practical data augmentation method mixup \cite{zhang2017mixup} in the computer vision field. We interpolate the smoothed representation with the original one-hot representation. 
Through interpolation, we can enlarge the probability of the original token, and the probabilities are still mostly distributed on the context-compatible words, as shown in the figure \ref{fig:ts_3}.

We combine the two stages as \textbf{text smoothing}: obtaining a smooth representation through MLM and interpolating to constrain the representation more controllable. To evaluate the effect of text smoothing, we perform experiments with low-resource settings on three classification benchmarks. In all experiments, text smoothing achieves better performance than other data augmentation methods. Further, we are pleased to find that text smoothing can be combined with other data augmentation methods to improve the tasks further. To the best of our knowledge, this is the first method to improve a variety of mainstream data augmentation methods.

\section{Related Work}
Various NLP data augmentation techniques have been proposed and they are mainly divided into two categories: one is to modify raw input directly, and the other interferes with the embedding \cite{miyato2016adversarial, zhu2019freelb}.
The most commonly used method to modify the raw input is the token replacement: randomly select tokens in a sentence and replace them with semantically similar tokens to synthesize a new sentence.
\cite{wei2019eda} directly uses the synonym table WordNet\cite{miller1998wordnet} for replacement.
\cite{kobayashi2018contextual} proposes contextual augmentation to predict the probability distribution of replacement tokens with two causal language models.
\cite{wu2019conditional} extends contextual augmentation with BERT's \cite{devlin2018bert} masked language modeling (MLM) to consider bi-directional context.
\cite{gao2019soft} softly augments a randomly chosen token in a sentence by replacing its one-hot representation with the distribution of the vocabulary provided by the causal language model in machine translation. Unlike \cite{gao2019soft}, we use MLM to generate smoothed representation, which considers the deep bi-directional context more adequately. And our method has better parallelism, which can efficiently obtain the smoothed representation of the entire sentence in one forward process. Moreover, we propose to constrain smoothed representation more controllable through interpolation for classification tasks.

\begin{lstlisting}[language=Python, 
                    caption=Codes to implement text smoothing in PyTorch, 
                    float=*]
    sentence = "My favorite fruit is pear ."
    lambd = 0.1 # interpolation hyperparameter
    mlm.train() # enable dropout, dynamically mask
    tensor_input = tokenizer(sentence, return_tensors="pt")
    onehot_repr = convert_to_onehot(**tensor_input)
    smoothed_repr = softmax(mlm(**tensor_input).logits[0])
    interpolated_repr = lambd * onehot_repr + (1 - lambd) * smoothed_repr
\end{lstlisting}

\section{Our Method}
\subsection{Smoothed Representation}
We use BERT as a representative example of MLM.
Given a downstream task dataset, namely $\mathcal{D} = \{t_i, p_i, s_i, l_i\}_{i=1}^N$, where $N$ is the number of instances, $t_i$ is the one-hot encoding of a text (a single sentence or a sentence pair), $p_i$ is the positional encoding of $t_i$, $s_i$ is the segment encoding of $t_i$ and $l_i$ is the label of this instance.
We feed the one-hot encoding $t_i$, positional encoding $p_i$ as well as the segment encoding $s_i$ into BERT, and fetch the output of the last layer of the transformer encoder in BERT, which is denoted as:

\begin{equation}
\overrightarrow{t_i} = \text{BERT}(t_i)
\label{bert-encoder}
\end{equation}

\noindent where $\overrightarrow{t_i} \in \mathcal{R}^{\text{seq\_len}, \text{emb\_size}}$ is a 2D dense vector in shape of [\text{sequence\_len}, \text{embedding\_size}].
We then multiply $\overrightarrow{t_i}$ with the word embedding matrix $W \in \mathcal{R}^{\text{vocab\_size,} \text{embed\_size}}$ in BERT, to get the MLM prediction results, which is defined as:
\begin{align}
\text{MLM}(t_i) = \text{softmax}(\overrightarrow{t_i}W^T)
\label{mlm-softmax}
\end{align}
\noindent
where each row in $\text{MLM}(t_i)$ is a probability distribution over the token vocabulary, representing the context-compatible token choices in that position of the input text learned by pre-trained BERT. 

\subsection{Mixup Strategy}
The mixup \cite{zhang2017mixup} is defined as:
\begin{align}
&\tilde{x}=\lambda x_{i}+(1-\lambda) x_{j} \\
&\tilde{y}=\lambda y_{i}+(1-\lambda) y_{j}
\end{align}
\noindent where $(x_i, y_i)$ and $(x_j, y_j)$ are two feature-target vectors drawn at random from the training data, and $\lambda \in [0,1]$.
In text smoothing, the one-hot representation and smoothed representation are derived from the same raw input, their lables are identical and the interpolation operation will not change the label. So the mixup operation can be simplified to:
\begin{align}
\widetilde{t_i} = \lambda \cdot t_i + (1-\lambda) \cdot \text{MLM}(t_i)
\label{text-smoothing}
\end{align}
\noindent where $t_i$ is the one-hot representation, $\text{MLM}(t_i)$ is the smoothed representation, $\widetilde{t_i}$ is the interpolated representation and $\lambda$ is the balance hyperparameter to control interpolation strength. 
In the downstream tasks, we use interpolated representation instead of the original one-hot representation as input.

\begin{table}[!t]
\centering
\small
\begin{tabular}{|l|c|c|c|}
    \hline
     & SST-2 & SNIPS & TREC \\
    \hline
    Train & 20 & 70 & 60 \\
    \hline
    Dev & 20 & 70 & 60  \\
    \hline
    Test & 1821 & 700 & 500 \\
    \hline
\end{tabular}
\caption{Data statistics in low-resource regime settings.}
\label{table_stat_exp1}
\vspace{-0.4cm}
\end{table}

\begin{table*}
\centering
\small
\begin{tabular}{|l|ccc|c|}
    \hline
    Method & SST-2 & SNIPS & TREC & Avg. \\
    \hline
    No Aug & 52.93 (5.01)  & 79.38 (3.20) & 48.56 (11.53)  & 60.29(6.58)\\
    \hline
    EDA & 53.82 (4.44)  & 85.78 (2.96) & 52.57 (10.49)  & 64.06(5.96)\\
    \hline
    BackTrans. & 57.45 (5.56)  & 86.45 (2.40) & 66.16 (8.52)  & 70.02(5.49)\\
    \hline
    CBERT & 57.36 (6.72) & 85.79 (3.46) & 64.33 (10.90) & 69.16(7.03) \\
    \hline
    BERTexpand & 56.34 (6.48) & 86.11 (2.70) & 65.33 (6.05)  & 69.26(5.08) \\
    \hline
    BERTprepend & 56.11 (6.33) & 86.77 (1.61) & 64.74 (9.61)  & 69.21(5.85) \\
    \hline
    GPT2context & 55.40 (6.71) & 86.59 (2.73) & 54.29 (10.12)  & 65.43(6.52)\\
    \hline
    BARTword & 57.97 (6.80) & 86.78 (2.59) & 63.73 (9.84)  & 69.49(6.41)\\
    \hline
    BARTspan & 57.68 (7.06) & 87.24 (1.39) & 67.30 (6.13) & 70.74(4.86)\\
    \hline
     Text smoothing  & \textbf{59.37(7.79)} & \textbf{88.85(1.49)} & \textbf{67.51(7.46)} & \textbf{71.91 (5.58)} \\
    \hline
\end{tabular}
\caption{Evaluating data augmentation methods on different datasets in a low-resource regime.}
\label{table_exp1_1}
\vspace{-0.2cm}
\end{table*}

\begin{table*}
\centering
\small
\begin{tabular}{|l|ccc|c|}
    \hline
    Method & SST-2 & SNIPS & TREC & Avg. \\
    \hline
    EDA & 59.66 (5.57) & 87.53 (2.31) & 55.95 (7.90) & 67.71 (5.26) \\
    + text smoothing & \textcolor{red}{\textbf{64.84(6.82)}} & \textbf{88.54(3.03)} & \textbf{67.68(9.70)} & \textbf{73.69(6.52)} \\
    \hline
    BackTrans. & 60.60 (7.40) & 86.04 (2.20) & 64.57 (7.48) & 70.40 (5.70) \\
    + text smoothing  &  \textbf{61.66(7.62)} & \textbf{88.72(1.99)} & \textbf{69.17(10.51)} & \textbf{73.19(6.7)} \\
    \hline
    CBERT & 60.10 (4.57) & 86.85 (2.06) & 63.56 (8.09) & 70.17 (4.91) \\
    + text smoothing  & \textbf{61.65(6.65)} & \textbf{88.18(2.85)} & \textbf{67.84(9.70)} & \textbf{72.56(6.4)} \\
    \hline
    BERTexpand & 59.85 (6.16) & 86.12 (2.45) & 62.67 (7.59) & 69.55 (5.40) \\
    + text smoothing  & \textbf{62.04(7.93)} & \textcolor{red}{\textbf{89.49(2.05)}} & \textbf{65.89(7.48)} & \textbf{72.47(5.82)} \\
    \hline
    BERTprepend & 60.28 (5.80) & 86.86 (2.46) & 65.20 (6.88) & 70.78 (5.05) \\
    + text smoothing  & \textbf{62.75(7.14)} & \textbf{88.04(1.92)} & \textbf{68.07(7.30)} & \textbf{72.95(5.45)} \\
    \hline
    GPT2context & 57.46 (4.96) & 84.10 (2.39) & 46.47 (12.80) & 62.68 (6.72) \\
    + text smoothing  & \textbf{60.66(6.72)} & \textbf{87.68(1.60)} & \textbf{59.13(11.33)} & \textbf{69.16(6.55)} \\
    \hline
    BARTword & 60.99(7.15) & 86.98(1.96) &  61.29(10.00) & 69.76(6.37) \\
    + text smoothing  & \textbf{62.67(7.40)} & \textbf{88.50(2.10)} & \textbf{67.75(6.50)} & \textbf{72.97(5.33)} \\
    \hline
    BARTspan & 63.42(5.58) & 87.34(2.17) & 62.47(8.11) & 71.08(5.29) \\
    + text smoothing  & \textbf{62.37(7.18)} & \textbf{89.06(2.18)} & \textcolor{red}{\textbf{70.89(6.81)}} & \textcolor{red}{\textbf{74.11(5.39)}} \\
    \hline
\end{tabular}
\caption{The effect of text smoothing combined with other data augmentation methods in low-resource regime.}
\label{table_exp1_2}
\vspace{-0.4cm}
\end{table*}

\section{Experiment}
\subsection{Baseline Approaches}
\textbf{EDA}\cite{wei2019eda} consists of four simple operations: synonym replacement, random insertion, random swap, and random deletion.\\
\textbf{Back Translation} \cite{shleifer2019low} translate a sentence to a temporary language (EN-DE) and then translate back the previously translated text into the source language (DE-EN).\\
\textbf{CBERT} \cite{wu2019conditional} masks some tokens and predicts their contextual substitutions with pre-trained BERT.\\
\textbf{BERTexpand, BERTprepend} \cite{kumar2020data} conditions BERT by prepending class labels to all examples of given class. ``expand" a the label to model vocabulary, while ``prepend" without.\\
\textbf{GPT2context} \cite{kumar2020data} provides a prompt to the pre-trained GPT model and keeping generating until the EOS token.\\
\textbf{BARTword, BARTspan} \cite{kumar2020data} conditions BART by prepending class labels to all examples of given class. BARTword masks a single word while BARTspan masks a continuous chunk.

\subsection{Experiment Setting}
Our experiment strictly follows the settings in the \cite{kumar2020data} paper on three text classification datasets downloaded from the links \footnote{SST-2 and TREC:\url{https://github.com/1024er/cbert_aug}, \\ SNIPS:\url{https://github.com/MiuLab/SlotGated-SLU/tree/master/data/snips}}.\\
\textbf{SST-2} \cite{socher2013recursive} is a movie reviews sentiment classification task with two labels.\\
\textbf{SNIPS} \cite{coucke2018snips} is a task of over 16,000 crowd-sourced queries distributed among 7 user intents of various complexity.\\
\textbf{TREC} \cite{li2002learning} contains six question types collected from 4,500 English questions.

We randomly subsample 10 examples per class for each experiment for both training and development set to simulate a low-resource regime.
Data statistics of the three datasets are shown in Table \ref{table_stat_exp1}. 
Following \cite{kumar2020data}, we replace numeric class labels with their text versions. 

We first compare the effects of text smoothing and baselines data augmentation methods on different datasets in a low-resource regime. Then we further explore the effect of combining text smoothing with each baseline method. Considering that the amount of data increases to 2 times after combination, we expand the data used in the baseline experiments to the same amount for the fairness of comparison. All experiments are repeated 15 times to account for stochasticity and results are reported as Mean (STD) accuracy on the full test set.

\subsection{Experimental Results}
As shown in Table\ref{table_exp1_1}, text smoothing brings the largest improvement to the model on the three datasets compared with other data augmentation methods. Compared with training without data augmentation, text smoothing achieves an average improvement of 11.62\% on the three datasets, which is significant. The previously best method is BARTspan, which is exceeded by Text smoothing with 1.17\% in average. 

Moreover, we are pleased to find that text smoothing can be well combined with various data augmentation methods, further improving the baseline data augmentation methods. As shown in Table\ref{table_exp1_2}, text smoothing can bring significant improvements of 5.98\%, 2.79\%, 2.39\%, 2.92\%, 2.17\%, 6.48\%, 3.21\%, 3.03\% to EDA, BackTrans, CBERT, BERTexpand, BERTprepend, GPT2context, BARTword, and BARTspan, respectively. To the best of our knowledge, this is the first method to improve a variety of mainstream data augmentation methods.

\section{Conclusoins}
This article proposes text smoothing, an effective data augmentation method, by converting sentences from their one-hot representations to controllable smoothing representations.
In the case of a low data regime, text smoothing is significantly better than various data augmentation methods. Furthermore, text smoothing can further be combined with various data augmentation methods to obtain better performance.

\bibliography{custom}
\bibliographystyle{acl_natbib}

\end{document}